# Exact Inference in Networks with Discrete Children of Continuous Parents

**Uri Lerner**
Computer Science Department
Stanford University
*uri@cs.stanford.edu*

**Eran Segal**
Computer Science Department
Stanford University
*eran@cs.stanford.edu*

**Daphne Koller**
Computer Science Department
Stanford University
*koller@cs.stanford.edu*

## Abstract

Many real life domains contain a mixture of discrete and continuous variables and can be modeled as *hybrid Bayesian Networks (BNs)*. An important subclass of hybrid BNs are *conditional linear Gaussian (CLG)* networks, where the conditional distribution of the continuous variables given an assignment to the discrete variables is a multivariate Gaussian. Lauritzen's extension to the clique tree algorithm can be used for exact inference in CLG networks. However, many domains include discrete variables that depend on continuous ones, and CLG networks do not allow such dependencies to be represented. In this paper, we propose the first "exact" inference algorithm for *augmented CLG* networks — CLG networks augmented by allowing discrete children of continuous parents. Our algorithm is based on Lauritzen's algorithm, and is exact in a similar sense: it computes the exact distributions over the discrete nodes, and the exact first and second moments of the continuous ones, *up to inaccuracies resulting from numerical integration used within the algorithm*. In the special case of *softmax* CPDs, we show that integration can often be done efficiently, and that using the first two moments leads to a particularly accurate approximation. We show empirically that our algorithm achieves substantially higher accuracy at lower cost than previous algorithms for this task.

## 1 Introduction

Bayesian Networks (BNs) provide a compact and natural representation for multivariate probability distributions in a wide variety of domains. Recently, there has been a growing interest in domains which contain both discrete and continuous variables, called *hybrid* domains. Examples of such domains include target tracking [1], where the continuous variables represent the state of one or more targets and the discrete variables might model the maneuver type; visual tracking (e.g., [13]), where the continuous variables represent the positions of various body parts of a person and the discrete variables the type of movement; and fault diagnosis [10], where a physical system contains continuous variables such as flows and pressures and discrete variables such as failure events.

The most commonly used type of hybrid BN is the *Conditional Linear Gaussian (CLG)* model. In CLGs, the distribution of the continuous variables is a linear function of their continuous parents, with Gaussian noise. Lauritzen [6, 7] showed that the standard clique tree algorithm can be extended to handle CLG networks, allowing the structure of the network to be exploited for inference, as in discrete BNs. Lauritzen's algorithm is "exact", in the sense that it computes the correct distribution over the discrete variables, and the correct first and second moments for the continuous ones. (It does not always compute the exact densities of the continuous variables, as these may be complex multi-modal distributions.)

Perhaps the main weakness of CLGs is that the graphical model does not allow discrete variables to have continuous parents, a dependency that arises in many domains. For example, consider a feedback control loop involving a thermostat, which controls the room temperature by turning on or off a heating device and a cooling system. The thermostat should be modeled using a discrete variable ("heating on", "cooling on", and "idle") which depends on the continuous variable representing the room temperature.

We can define a class of *augmented CLG* networks, which uses CLG dependencies for the continuous variables, but also allows dependencies of discrete variables on continuous parents [4]. The conditional probability distributions (CPDs) of these nodes are often modeled as *softmax* functions, which include as a special case a "soft" threshold function for a continuous parent (i.e., a noisy indicator whether the value of the continuous parent is greater than some constant). There are many domains that can be modeled very naturally using augmented CLG networks, including our thermostat example above.

Unfortunately, there is no exact inference algorithm known for augmented CLG networks. One can always resort to the use of approximate inference, such as discretization (e.g., [5]) or sampling (either *Likelihood Weighting* [15] or *Gibbs Sampling* [12]), but these approaches have some serious limitations. It is often hard to find a good discretization: Sometimes any reasonable discretization demands too fine a resolution, and often requires the handling of intractable intermediate factors (especially in high dimensions). The convergence of sampling algorithms can be quite slow, and is very sensitive to the network parameters and the configuration of the evidence. Murphy [11] proposed a variational approximation for a class of augmented CLG networks, based on the observation that a Gaussian can be a good approximation to the product of a Gaussian and a softmax. Wiegerinck [16] showed how this approach can be adapted to deal with multi-modal distributions. However, this approach is currently limited to binary discrete variables and softmax CPDs. More importantly, it does not provide any performance guarantees on the quality of the resulting approximation.

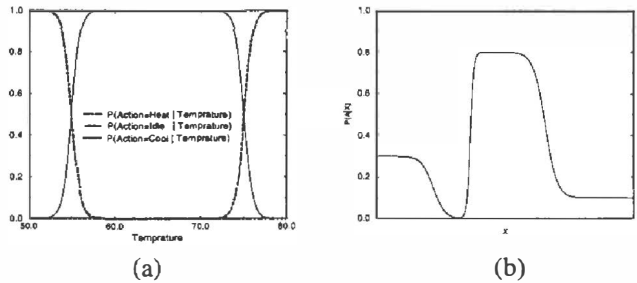

(a)          (b)

Figure 1: Examples of softmax CPDs: (a) The thermostat (b) Multi-transition softmax

In this paper we propose the first "exact" algorithm for augmented CLG networks. Our algorithm is based on the following simple, yet powerful, idea. Consider the case where the discrete children are modeled with softmax CPDs. As in [11], we approximate the product of a Gaussian and a softmax as a Gaussian, but rather than using a variational approach, we find the approximation directly using numerical integration. We embed this idea within the general framework of Lauritzen's algorithm for CLG networks, leading to a simple algorithm, which is roughly comparable in its complexity to Lauritzen's algorithm.

We prove that our algorithm is exact, in a sense that is analogous to Lauritzen's algorithm: It computes the exact distributions over the discrete nodes, and the exact first and second moments of the continuous ones, *up to inaccuracies resulting from numerical integration used within the algorithm*. We also show empirically that it achieves extremely high accuracy for "reasonable" numerical integration schemes, leading to results that are significantly better than current approximate inference algorithms.

## 2 Hybrid Bayesian Networks

A hybrid BN represents a probability distribution over a set of random variables where some are discrete and some are continuous. We denote discrete variables with letters from the beginning of the alphabet ($A$, $B$, $C$, and $D$) and continuous ones with letters from the end ($X$, $Y$, and $Z$).

Sets of variables are denoted with boldface (e.g., $\boldsymbol{X}$).

Bayesian networks consist of two parts: a qualitative component given by a directed acyclic graph whose nodes correspond to the random variables and a quantitative component given by a set of *conditional probability distributions (CPDs)* which define the conditional distribution of every node given its parents in the graph. In principle, there is no restriction on the form of the CPDs. However, if we want to perform efficient inference on hybrid BNs, some restrictions are necessary.

The most widely used subclass of hybrid BNs is the *conditional linear Gaussian (CLG)* model. Let $X$ be a continuous node, $\boldsymbol{A}$ be its discrete parents and $Y_1, \ldots, Y_k$ its continuous parents. The CLG CPD defines $X$ to be a linear Gaussian function of the $Y_i$'s, with a different set of parameters for the linear Gaussian for every instantiation of $\boldsymbol{A}$:

$$p(X \mid \boldsymbol{a}, \boldsymbol{y}) = N(w_{\boldsymbol{a},0} + \sum_{i=1}^{k} w_{\boldsymbol{a},i} y_i; \sigma_{\boldsymbol{a}}^2).$$

The CLG model does not allow discrete children of continuous parents. This model has the appealing property that it defines a conditional Gaussian joint distribution: for any assignment to the discrete variables, the distribution over the continuous variables is a multivariate Gaussian. The reason is that, given such an assignment, the CPDs of the continuous nodes reduce to simple linear Gaussians, inducing a multivariate Gaussian.

As discussed in the introduction, the inability of the CLG model to represent the dependence of discrete nodes on continuous ones severely restricts their range of applicability. In this paper, we define a class of *augmented* CLGs, where the CPDs of the continuous nodes are CLGs, as discussed above, but where we also allow discrete nodes to depend on continuous parents. We use *continuous-discrete (CD)* to refer to such CPDs.

There are many possible functional forms that can be used to represent CD CPDs. One of the most useful is a *softmax* or *logistic* function. Let $A$ be a discrete node with the possible values $a_1, \ldots, a_m$, and let $Y_1, \ldots, Y_k$ be its parents. We define:

$$P(A = a_i \mid y_1, \ldots, y_k) = \frac{\exp\left(b^i + \sum_{l=1}^{k} w_l^i y_l\right)}{\sum_{j=1}^{m} \exp\left(b^j + \sum_{l=1}^{k} w_l^j y_l\right)}.$$

The case where $A$ also has discrete parents is modeled as in CLGs: we define a different softmax function for every combination of the discrete parents. It is possible to eliminate one of the linear combinations by dividing both numerator and denominator by $\exp\left(b^i + \sum_{l=1}^{k} w_l^i y_l\right)$, leaving us with $m-1$ sets of parameters. In particular, when $A$ is binary, this new form simplifies to a standard sigmoid function:

$$P(A = a_1 \mid y_1, \ldots, y_k) = \frac{1}{1 + \exp\left(b + \sum_{l=1}^{k} w_l y_l\right)}.$$

Fig. 1(a) shows a softmax CPD for our thermostat example from the introduction. The CPD parameters control the location and the slope of the transitions. It is possible to generalize the softmax functions to express fairly complex distributions, as in Fig. 1(b); see [4] for discussion.

## 3 Inference in CLGs

We present a brief review of Lauritzen's algorithm for inference in CLG networks, on which our algorithm is based. The algorithm has two versions: the original one [6] and an improvement to it [7]. Both versions are based on the *clique tree algorithm* [8]. The clique tree algorithm begins by transforming the BN into a *clique tree*. The first step is to generate the *moralized graph*, where the parents of each node are connected directly, and all edges are undirected. The moralized graph is then transformed into a clique tree using a process called *triangulation* (see [8] for details). In the clique tree, each node (also called a clique) is associated with a data structure, called a *potential*, that can represent a function over the possible values of the variables in the clique (in general, these data structures are called *factors*). In a purely discrete BN, the factor is typically a table with one entry (a number) for each assignment to the variables in the clique. In the message passing phase of the algorithm, factors are passed between neighboring cliques. At the end of this phase, every clique potential contains the correct marginal distribution over the clique variables.

Several issues arise when extending the clique tree algorithm to CLG networks. Most obviously, the factors in the cliques and the messages are functions over both discrete and continuous variables and cannot be represented by tables as in the discrete case. Lauritzen's algorithm deals with this issue by defining a factor as a table which specifies a continuous function for every instantiation of the discrete variables. The two versions of the algorithm use these continuous functions in different ways. In the original version, the functions are treated as *canonical forms*, which can represent any function of the form $P(x; g, h, K) = \exp[g + x'h + x'Kx]$, where $g$ is a constant, $h$ is a vector and $K$ is a full-rank square matrix. Note that a multivariate Gaussian, whose density is $\frac{1}{\sqrt{2\pi|\Sigma|}} \exp\left(-(x-\mu)^t \Sigma^{-1}(x-\mu)/2\right)$, is a special case of this form (see [6] for formulae to convert from a multivariate Gaussian to canonical form). However, not every function representable in canonical form is a multivariate Gaussian (e.g., $\exp(x^2)$). In fact, canonical forms can represent functions which are *not* probability distributions: They do not necessarily have a finite integral and their moments may not be defined. In particular, CLG CPDs, which represent a conditional rather than a joint distribution, are representable in canonical form but not as a Gaussian. In the new version of the algorithm, the factors represent conditional Gaussians, i.e., they represent a conditional distribution of a subset of the variables given the rest.

The clique tree algorithm manipulates factors in various ways, such as multiplying, dividing and marginalizing. Lauritzen shows how all these operations can be carried out exactly in both versions of the algorithm, with the notable exception of summing out a discrete variable. For example, consider a factor over the variables $A$ and $X$ (where $A$ is discrete and $X$ is continuous) and assume we need to compute its marginal over $X$ in order to send a message to a neighboring clique (i.e., sum out the variable $A$). Since the message contains only one Gaussian, we need to *collapse* the two Gaussian components in the original mixture, while maintaining the correct first and second moments. While we can collapse Gaussians using their moments, the operation is not defined for a general canonical form or for a conditional Gaussian in which the moments may not be well defined. Thus, we must ensure that when the message passing algorithm calls for collapsing, our factors will represent Gaussians.

To ensure this property, Lauritzen's algorithm imposes certain constraints on the form of the clique tree. These lead to the notion of *strong triangulation*. While the exact details are not important for the purposes of this paper (see [6, 7]), one of the implications of strong triangulation is important for our analysis. Define a *continuous connected component* as a set of continuous variables $X$ such that every two variables $X_1, X_2 \in X$ are connected in the moralized graph via a chain consisting only of continuous variables. We define $\text{DN}(X)$, the *discrete neighbors of* $X$, as the set of discrete variables that are adjacent to some variable in $X$ in the moralized graph. Strong triangulation implies that all the variables in $\text{DN}(X)$ necessarily appear together in some clique in the tree. The intuition for this requirement is that the distribution over $X$ is a mixture of Gaussians with one mixture component for every assignment to $\text{DN}(X)$; hence, we must consider all the combinations of the discrete neighbors together.

The cost of Lauritzen's algorithm is polynomial in the size of the factors in the cliques. This size grows exponentially with the number of discrete variables in the clique, and quadratically with the number of continuous variables in the clique. Thus, the strong triangulation property, although unavoidable, is a major computational limitation of Lauritzen's algorithm. (See [9] for further discussion.)

## 4 Inference in Augmented CLGs

We now extend Lauritzen's algorithm to the class of augmented CLG networks defined in Section 2. We present our algorithm in the context of the original version of Lauritzen's algorithm and later show how it can be easily adapted for the modified version.

We first motivate our algorithm with a simple example. Consider the network $X \to A$, where $X$ has a Gaussian distribution given by $P(X) = N(\mu, \sigma)$ and the CPD of $A$ is a softmax given by $P(A = 1 \mid X = x) = 1/(1 + e^{ax+b})$. The clique tree has a single clique $(X, A)$, whose factor should contain the product of these two CPDs. Thus, it should contain two continuous functions — $P(x)P(A = 1 \mid x)$ and $P(x)P(A = 0 \mid x)$ — each of which is a product of a Gaussian and a sigmoid.

Our algorithm is based on the observation [11] that the product of a Gaussian and a sigmoid can be approximated quite well by a Gaussian distribution. We can compute the best Gaussian approximation to this function by computing the marginal distribution of $A$ and of the first and second moments of $X$ from the joint distribution:

$$\begin{aligned} P(A = a) &= \int_{-\infty}^{\infty} P(A = a \mid x) P(x) dx \\ E[X \mid A = a] &= \int_{-\infty}^{\infty} x P(x \mid A = a) dx \\ &= \frac{1}{P(A=a)} \int_{-\infty}^{\infty} x P(A = a \mid x) P(x) dx \end{aligned} \quad (1)$$

Similarly, we compute the second moment $E[X^2 \mid A = a]$.

This basic idea leads us to the following outline for an algorithm. We roughly follow Lauritzen's algorithm, diverging only in cases where a clique contains CD CPDs; in this case, we approximate its factor as a mixture of Gaussians, where the mixture has one Gaussian — with the correct first and second moments — for each instantiation of the discrete variables (no matter their configuration). In the remainder of this section, we "fill in" the details of this algorithm, addressing the subtleties that arise.

### 4.1 The algorithm

The first difficulty arises from the observation that the equations in (1) compute expectations relative to $P(x)$: To evaluate these expressions at a clique, we must have a probability distribution over $X$ at that clique. Unfortunately, the message passing algorithm does not guarantee that these distributions are available initially. Consider the network $X \rightarrow Y \rightarrow A$. The clique tree for this network consists of two cliques: $(X, Y)$ and $(Y, A)$. In Lauritzen's algorithm, the clique tree is initialized by incorporating all CPDs into their corresponding cliques. In our case, we should incorporate $P(A \mid Y)$ into the clique $(Y, A)$ by computing the relevant expectations. However, at the initialization phase, the message passing has not yet been performed. As such, $Y$ is given in canonical form and does not yet represent a Gaussian distribution, preventing us from performing this integration; thus, we cannot multiply the CPD $P(A \mid Y)$ into the clique at this stage.

We address this problem by introducing a preprocessing phase, which serves to guarantee that all cliques contain an *integrable distribution* — a Gaussian distribution relative to which we can compute the relevant expectations, rather than a non-Gaussian canonical form. To do so, we build the standard clique tree for our BN, but do not initialize the clique potentials. We then insert all the CPDs except for the CD CPDs. The resulting network is equivalent to a CLG network, so we can calibrate it using Lauritzen's algorithm, resulting in probability distributions in each clique. Finally, we insert the remaining CD CPDs and re-calibrate the tree. Note that the cliques in our tree were designed to accomodate this insertion operation. Since we now have integrable distributions, we can perform the approximation.

Our solution to this problem raises the following question: Can we use the prior distribution over the CLG component as our integration distribution? Unfortunately, there are several reasons why the use of this distribution is an approximation which can lead to errors. We now discuss each of these, and show how to correct them.

The first difficulty is that our prior distribution is computed before incorporating the evidence. Consider, for example, the network $X \rightarrow Y \rightarrow A$, and assume that $X$ is observed. The minimal cliques are $(X, Y)$ and $(Y, A)$. Following our current algorithm, we would insert the CPD for $P(A \mid Y)$ and calibrate the tree, approximating it as a CLG network. If we now enter the evidence observed for $X$, we would be incorporating it into an approximate distribution rather than the true one, potentially leading to sub-optimal approximations. Fig. 2(a) shows an example of this phenomenon, where the approximation obtained by first integrating the CD CPD and then conditioning on our evidence is a sub-optimal approximation. The optimal approximation uses the posterior over $Y$ directly as our integration distribution. Our solution to this problem is straightforward: We not only ensure that each clique has a Gaussian distribution in it, we ensure that it has the posterior Gaussian. Thus, we incorporate the evidence and propagate it before entering the softmax CPDs.

A more subtle problem with our choice of integration distribution relates to the use of collapsing within Lauritzen's algorithm. Consider a network $A \rightarrow X \rightarrow Y \rightarrow B$, and assume that the clique tree has the cliques $(X, Y, B)$ and $(A, X, B)$ (note that the tree $(A, X), (X, Y), (Y, B)$ is inconsistent with strong triangulation). According to our current algorithm, we calibrate the clique tree with the CPDs for $A$, $X$, and $Y$, and then insert the CPD $P(B \mid Y)$ into the clique $(X, Y, B)$. However, the distribution in this clique is not the correct prior distribution over $Y$. The correct prior distribution of $Y$ has two modes (one for every value of $A$); but, as $A$ does not appear in the clique, Lauritzen's algorithm collapses the two modes into a single Gaussian, losing its bimodal nature. Although Lauritzen shows that this approximation can be used without introducing new errors if the functions are linear, the CPD $P(B \mid Y)$ is not linear, and we may not get the best approximation for the first two moments. Fig. 2(b) shows an example, where our approximation is worse when we use the collapsed distribution for $Y$ as the integration distribution.

Once again, the solution is to enter the CD CPD into a clique where the integration distribution over the continuous parents is correct. Let $X$ be the continuous parents of the CD CPD, let $Y \supseteq X$ be the variables of their continuous connected component (as defined in Section 3), and let $A = DN(Y)$. As we discussed, $X$ will have one mode for every assignment to $A$. Hence, if we want to represent the exact multi-modal distribution for $X$, it is necessary and sufficient to have a clique containing the variables in both $X$ and $A$. Of course, this requirement could result in a larger clique tree; however, the overhead is not large. As we discussed earlier, $A$ must be in some clique in the optimal tree. Thus, at worst, we only add some continuous variables to some of the cliques. Since the represntation of canonical forms (and multivariate Gaussians) is quadratic in the number of variables, the size of the tree can only grow by a polynomial factor at worst.

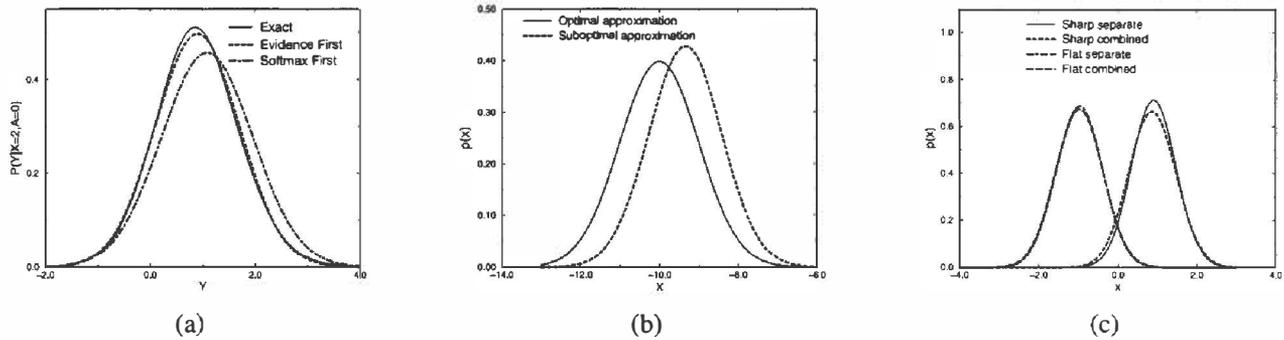

Figure 2: (a) Incorporating softmax CPDs before and after the evidence. (b) Error introduced if the discrete neighbors are not in the same clique as the sigmoid CPDs. (c) Error introduced if the sigmoid CPDs are entered separately.

Note that this modification to the clique tree is necessary only if we want to guarantee the optimal approximation. The algorithm remains coherent if we use an approximate integration distribution, only the quality of our approximation can degrade. Therefore, we can use a clique tree where the clique that contains $X$ contains only some subset of the variables in $A$.

The final problem arises when there is more than one CD CPD. Most simply, we can insert each CD CPDs sequentially. We insert each CD CPD, approximate the resulting joint distribution as a mixture of Gaussians, and proceed to use that mixture as the basis for inserting the next one. The obvious problem with this approach is that the integration distribution used for the CD CPDs inserted later is only an approximation to the correct non-Gaussian distribution resulting from the insertion of the earlier CD CPDs. The solution to the problem is to integrate all the CD CPDs in the same continuous connected component in one operation. In Fig. 2(c), we show the difference on the network $A \leftarrow X \rightarrow Y \rightarrow B$. We tried inserting both softmax CPDs into the clique containing $X$ and $Y$ together, and separately. We experimented both with step-like transitions ("sharp" sigmoids), and smoother transitions. The latter allows for a better approximation as a Gaussian, and therefore less error by doing the approximation step by step. This difference is clearly manifested in the figure.

While the idea of joint integration seems expensive, we note that the relevant CPDs must already be in the same clique (with their discrete neighbors), so we do not increase the size of the tree. However, we do pay the price of com-

Construct a strongly triangulated clique tree such that for every
    maximal connected component $X_i$ there exists a clique $C_i$
    such that $C_i$ contains $X_i$ and its discrete neighbors
Insert all CPDs except for softmax CPDs
Calibrate the tree using Lauritzen's algorithm
Insert the (continuous and discrete) evidence and re-calibrate
Instantiate the CD CPDs with the continuous evidence
**for** each maximal connected component $X_i$
    Find all softmax CPDs $S_1, ..., S_n$ that can go into $C_i$
    Insert $S_1, ..., S_n$ into $C_i$ using multi-dimension integration
Re-calibrate the tree
Return the distribution over $Q$

Figure 3: Outline of full algorithm

puting integrals in higher dimensions. We can reduce this cost by integrating only some of the CD CPDs together. This scheme induces a spectrum of approximations, with a tradeoff between complexity and accuracy: If doing the high dimensional integration is intractable, we can approximate it either by inserting the CPDs separately or by using a more efficient and less accurate integration method.

The full algorithm for inference in augmented CLG networks is presented in Fig. 3. We are given a hybrid Bayesian network $B$, evidence $e$ and a query $Q$ and wish to compute $P(Q, e)$. Note that the CD CPDs should be integrated together to achieve the best approximation but can also be integrated separately as discussed above.

### 4.2 In-clique integration

Having defined the overall structure of the algorithm, it remains only to discuss the integration process within each clique. There is a wide range of numerical integration methods that can be applied in our setting. We focus our attention on one that seems particularly suitable in our framework — *Gaussian quadrature integration* [2]. Gaussian quadrature approximates a general integral as follows:

$$\int_a^b W(x)f(x)dx \approx \sum_{j=1}^N w_j f(x_j)$$

Where $W(x)$ is a known function and the points $x_j$ and weights $w_j$ are selected such that the integral would be exact if $f(x)$ is a polynomial of degree $2N - 1$ or less. Quadrature is particularly well suited in our setting, since we can set $W(x)$ to be a Gaussian, for which good points and weights are known (see [14] for code to generate the points and weights). The main drawback of this method is that the size of the grid grows exponentially with the dimension of the integral, which seems to be a real problem, as the cliques in our algorithm often contain many continuous variables (particularly if we keep a clique which contains an entire continuous connected component). Fortunately, we can use the properties of our augmented CLG networks to significantly reduce the computational burden.

Our first observation exploits the fact that we are dealing with Gaussian distributions. Assume that the continuous variables can be partitioned into the sets $Y$ and $Z$ where only variables from $Y$ appear in CD CPDs. Recall that we can represent the multivariate Gaussian over the variables $Y, Z$ as a linear Gaussian network with the structure $Y \to Z$ (i.e., there are no edges from a variable in $Z$ to a variable in $Y$). The CD CPD changes the distribution over $Y$, and since any variable in $Z$ depends linearly on the variables in $Y$, having the first two moments for $Y$ is enough to infer the first two moments for $Z$ without any further numerical integration. Thus, to incorporate the variable $A$ into the CPD, the required integration dimension is exactly the number of continuous parents of $A$.

Surprisingly, we can substantially improve even on this idea, in the case where the CPD of $A$ is a softmax. The softmax for a node $A$ is a soft threshold defined via a set of linear functions $f_a$ over the continuous parents $Y$ of $A$; we have one function for each value $a$ of $A$, although we can eliminate one of them as discussed in Section 2. We can now define a set of new variables $Z_a$ which are a (deterministic) linear function of the variables $Y$: $Z_a = f_a(Y)$. We can then reinterpret the softmax as a CPD whose parents are the variables $Z_a$. (More generally, we can use any set of variables $Z$ which are linear combinations of $Y$ such that every $f_a$ can be presented as a linear combination of $Z$.) Note that, as the $Z_a$'s are linear functions of the parents $Y$, a Gaussian distribution over $Y$ induces a Gaussian distribution over the $Z_a$'s. We can use the distribution over the $Z_a$'s as our integration distribution, and then propagate the result to the actual parents $Y$ using the linearity between the $Y$'s and the $Z_a$'s. The dimension of the integrals we have to perform is at most $|A| - 1$, where $|A|$ is the number of values of $A$. When dealing with binary variables, this approach can result in dramatic savings.

Of course, one can still construct networks where the integration dimension is very large: networks where the discrete variable $A$ has many values and continuous parents, or where there are many CD CPDs that all need to be integrated into the same clique. When forced to deal with these cases, we have several choices. We can resort to some approximation, e.g., inserting the CD CPDs one at a time rather than all at once, thereby losing the optimality guarantees of our algorithm. Alternatively, we can resign ourselves to the use of other numerical integration methods, such as Monte Carlo integration, that scales better with the dimensionality of the (continuous) space. Another possibility is the use of adaptive integration methods, described in [2]. Here we assume that we have available an integration procedure that also outputs an error estimate. We then use this procedure to adaptively focus the computational efforts to regions which produce larger estimated errors (in our case we would use most of our resources in areas where the sigmoids transition sharply between 0 and 1).

### 4.3 Using Lauritzen's modified algorithm

Lauritzen's original algorithm suffers from some well known numerical instabilities which our extension to it would inherit. Lauritzen's modified algorithm improves upon the original one by providing better numerical stability, and also by dealing with deterministic CPDs. Unlike the original version, it maintains conditional distributions in the cliques, except for the strong root, in which a multivariate Gaussian is kept. To enjoy the benefit of numerical stability, we can adapt our algorithm to work with Lauritzen's modified algorithm. All we need to do is to ensure that CD CPDs are inserted to a clique which is a strong root, guaranteeing it has an integrable distribution. If the strong root does not naturally contain the continuous variables from the CD CPDs, we can redesign the clique tree to ensure that this property holds. (This process can be accomplished using PUSH operations [7].) The only possible consequence is the addition of continuous variables to some cliques. Therefore, the added complexity is quadratic in the number of continuous variables in the worst case.

## 5 Analysis

We now show that our algorithm is "exact", up to errors caused by numerical integration. We use "exact" in the same sense used in Lauritzen's algorithm: It computes the correct distribution over the discrete nodes, and the correct first and second moments for the continuous ones.

**Theorem 1** *Let $Q$ be a query such that $Q \subseteq C$ where $C$ is some clique in the tree and let $e$ be some evidence. The above algorithm computes a distribution $P(Q, e)$ which is exact for discrete variables in $Q$ and has the correct first two moments for continuous variables in $Q$, up to inaccuracies caused by numerical integration.*

**Proof:** We start by showing that the algorithm is exact when the moralized graph contains one continuous connected component and the clique tree has just one node. The algorithm has three steps, and we analyze the result of applying each one of them. The first step involves inserting discrete and CLG CPDs into the clique tree. Since all the variables are in one clique (hence there is no need for collapsing) we get that the clique potential represents the exact product of the CPDs that were inserted. Note that the product of the clique potential and the CD CPDs is the exact prior distribution.

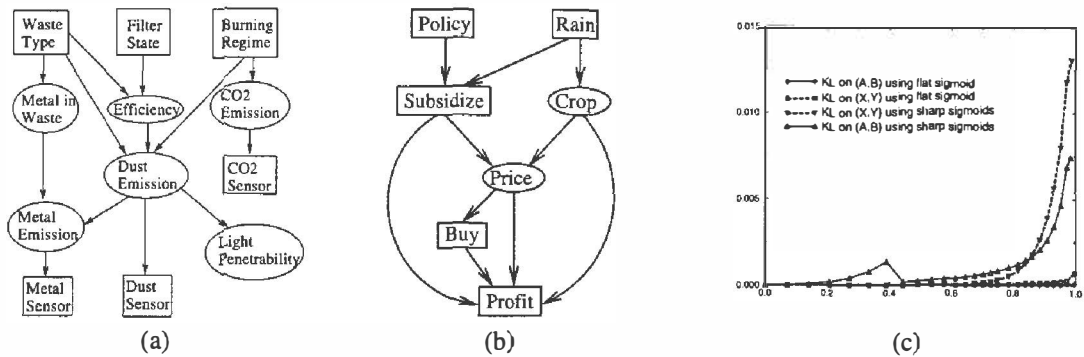

Figure 4: (a) The Emission network. Ovals are continuous variables and rectangles are discrete. (b) The extended crop network. (c) Error caused by inserting CD CPDs separately.

The next step involves incorporating the evidence. The discrete evidence can be viewed as extra factors that are multiplied into the tree (the entry in the factor corresponding to the evidence is 1 and the rest are 0). Multiplying these factors into the clique tree is an exact operation, and does not introduce any inaccuracies. We now consider the continuous evidence. Setting the evidence is equivalent to setting the relevant values in every function of the product of the tree and the CD CPDs. In the clique potential this can be done using Lauritzen's standard algorithm. In the CD CPDs we are guaranteed that any continuous evidence variables that appear in the CPD must be parents; Thus, we can simply set their value and get a new conditional distribution not involving them, which is either a CD CPD or just a discrete factor.

The last step is the insertion of the CD CPDs into the clique potential. Had we been able to represent the answer exactly, then the clique potential would have been the exact posterior distribution. However, we approximate each of the continuous components in the potential as a Gaussian, computed in a way that guarantees it has the correct first two moments (up to numerical integration inaccuracies).

We now remove the various assumptions. The clique $C$ where the CD CPDs are entered need not contain discrete variables which are not discrete neighbors of the continuous connected component, since it can represent the exact distribution over the continuous variables without them. It also need not contain continuous variables which are not involved in any CD CPD. The key insight here is that after incorporating all the non-linear functions into $C$, the rest of the cliques have linear functions. Therefore, in order to find the correct first two moments, it suffices to use message passing, where each message has the correct first two moments. (This claim is similar to the proof of correctness for Lauritzen's algorithm.) Finally, assume that the moralized graph has more than one continuous connected component. In this case, strong triangulation guarantees that no clique contains continuous variables from two different continuous connected components. Thus, the messages passed between different continuous connected components are discrete factors which do not change the conditional distribution of the continuous variables given their discrete neighbors. ∎

## 6 Experimental Results

We tested various aspects of our algorithm and compared it to other approaches. As we discussed, Gaussian quadrature does not scale up well in high dimensions. To demonstrate the effects of integrating in high dimensions we tested our algorithm on networks where the continuous variables form a chain $X_1 \to \cdots \to X_n$, with each $X_i$ a parent of a discrete variable $A$. We varied $n$, thus simulating integration problems in different dimensions. Fig. 5(a) and (b) show the result of performing the integration in one and eight dimensions using both Gaussian quadrature and Monte Carlo integration. In one dimension, Gaussian quadrature is extremely efficient, achieving good accuracy with as few as 5 integration points. In eight dimensions, Gaussian quadrature needs many more points to achieve a similar accuracy. Note that we need at least two points in every dimension, for a total of at least 256 points in 8 dimensions. As we discussed, it is possible to dramatically speed up the integration by taking advantage of the sigmoid function, representing the linear combinations of the eight parents as one variable; indeed, we get a reduction of approximately three orders of magnitudes in the number of points required. Note that this case still converges more slowly than simple integration in one dimension. The reason is that the variance of the dummy parent is larger than the variance we used in one dimension (we further discuss this issue in Section 7). Finally, we can see that although Monte Carlo integration converges quite slowly, it is almost unaffected by the higher dimension, and can be used in cases where we have no choice but to integrate in high dimension.

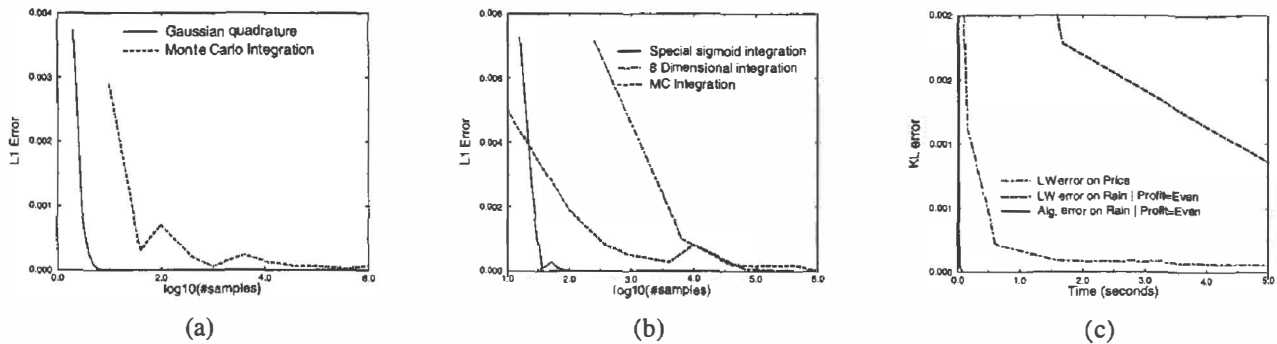

Figure 5: (a) Integration in one dimension. (b) Integration in eight dimensions. (c) Comparison with Likelihood Weighting

We then considered the effect of inserting the CD CPDs together versus one at a time. We experimented with the network $A \to X \to Y \to B$ used in Fig. 2(c). In Fig. 4(c) we show the KL-distance, for both the discrete and the continuous variables, between the distributions obtained by inserting the softmax CPDs together versus one at a time. The error depends on the strength of the correlation between $X$ and $Y$, so we considered different correlation values ranging from 0 to 1. We also changed the parameters of the softmax CPDs, allowing for smooth transitions in one case (a flat sigmoid) and step-like transitions in another case. We see that the error is larger when there is a strong correlation between variables influenced by different CD CPDs, and when the sigmoids are sharp, making the Gaussian approximation worse. Interestingly, however, the accuracy is very high in all cases; Thus, even if we insert CD CPDs separately, with the associated computational benefits, we often get very accurate results.

We compared the accuracy of our algorithm to that of others on more realistic examples that have been considered by other researchers. We first tested our algorithm on the Crop network presented by Murphy in [11]. We compared our results with the results of Murphy and with those of Gibbs sampling using BUGS [3]. It turns out that Murphy's variational algorithm performs quite poorly when the posterior distribution is multi-modal, achieving L1-errors over the binary discrete variables of 0.28–0.38. On the other hand, both our algorithm and BUGS performed very well on this simple network, giving the correct result almost instantaneously. We also note that Wiegerinck [16] reports good results for this network using his variational approach.

To test the algorithm on a larger network, we used the *Emission* network described in [6], which models the emission of heavy metals from a waste incinerator. The original network is a CLG. We augmented it with three extra discrete binary variables as shown in Fig. 4(a). The additional variables correspond to various emission sensors and each has a CD CPD: a dust sensor ($w = 1, b = -3$), a $CO_2$ sensor ($w = 3, b = 6$) and a metal emission sensor ($w = 2, b = -5.6$).

We experimented with various queries using our algorithm and compared it to a few runs of Gibbs sampling using BUGS. We found that BUGS seemed unstable and produced results which differed significantly from ours. As an example, we queried for the distribution over the emission of dust after setting both the metal emission sensor and the $CO_2$ sensor to *High*. Our algorithm returned a mean of 3.419 and a variance of 1.007. BUGS converged after about 500,000 samples to a mean of 3.31 and a variance of 0.31, which did not change substantially even after 1,000,000 samples. To understand the discrepancy, we used likelihood weighting on the same query. After 500,000 samples, the estimated mean was 3.418 and the estimated variance was 0.999 which agree quite closely with the results produced by our algorithm.[1] We note that our algorithm achieved these results with only 3 quadrature points per dimension, and with the highest integration dimension being 2. Hence, our algorithm was almost instantaneous, and was much faster than both BUGS and likelihood weighting.

As a final example, we tried our algorithm with a network containing non-softmax CD CPDs. We augmented the crop network (see Fig. 4(b)) with three more variables. One of them is the *Profit* variable, which depends on a product of the *Crop* and *Price* variables. The parameters of the extended network appear in Appendix A. Having experienced problems using BUGS, we compared our results to likelihood weighting. We tested two scenarios, one without evidence, and one with the evidence *Profit=Even*, and compared the accuracy of both algorithms on various queries. We used numerical integration with 150 points per dimension as our ground truth. We ran LW and our algorithm for the same amount of time, and then measured the KL-distance between the "ground truth" and the results. For LW, we averaged over 10–500 runs (we used more runs for smaller number of samples where the variance is bigger). Fig. 5(c) shows the results for the KL-error for *Price* with no evidence and for *Rain* given *Profit=Even*. Only three lines are visible: the KL-error of our algorithm in the case of no evidence is too close to zero to be visible in the graph. It is clear that our algorithm converges much faster than LW, especially when we have evidence. This is to be expected, as it is well known that sampling methods can take a lot of time to converge, and the performance of LW deteriorates when there is evidence.

---

[1] We further investigated this discrepancy and discovered that BUGS also returns answers that disagree with those appearing in [6] even for the original *Emission* network without the CD CPDs. For example the standard deviation for *DustEmission* converged to 0.85 instead of 0.77 (the mean was correct).

## 7 Discussion

In this paper, we presented the first exact inference algorithm for augmented CLG networks. We use numerical integration to compute the first two moments of every mixture component and thus approximate it as a multivariate Gaussian. We show how this approach can be incorporated into Lauritzen's clique tree algorithm (both the original and the modified version), which enables us to take advantage of the properties of the network to speed up the computation. In particular, our algorithm exploits both the linearity of the CLG part of the network, and the properties of the softmax CPDs (when applicable), to reduce the dimension of the integration. We proved that our algorithm produces the correct distribution over the discrete variables, and the correct first two moments of the continuous variables, up to inaccuracies resulting from numerical integration. Thus, it gives the best approximation within our expressive power.

Our algorithm is not restricted to any special class of the conditional distributions in the CD CPDs — we can always compute the correct first two moments, resulting in a Gaussian approximation. However, the quality of the Gaussian approximation varies for different classes of conditional distributions. In the common case of softmax CPDs, the Gaussian approximation is often an excellent approximation to the true posterior distribution.

As our algorithm relies heavily on numerical integration, its performance is directly related to the quality and efficiency of the numerical integration procedure. The Gaussian quadrature method works particularly well in many networks but runs into problems in high dimensions and when the sigmoids are sharp relative to the variance of the Gaussian (i.e., they resemble a step function rather than a smooth transition). The reason for this problem is that Gaussian quadrature tries to find a set of points which optimizes the performance for functions which are polynomials. Since the smoother the function is, the better its approximation as a low degree polynomial, Gaussian quadrature is more accurate with smooth sigmoids. We point out that, in the case of sharp sigmoids, a Gaussian approximation of the posterior can be quite bad and one may not want to use it regardless of the numerical integration accuracy. In any case, when Gaussian quadrature is not well suited for the problem, we can use other integration methods such as adaptive integration and Monte Carlo methods.

Another problem for the algorithm relates to unlikely discrete evidence (unlikely continuous evidence is not problematic), because even slight errors in the distribution $P(Q, e)$ are magnified by the renormalization process. We can reduce the effect of unlikely evidence by a simple two-step process: we run our algorithm to obtain a first estimate of the posterior over the discrete variables and then rerun it, allocating more of our resources to the mixture components that are more likely in the posterior distribution. We plan to test this approach in future work.

Existing methods for inference in augmented CLG networks can be divided into three classes: discretization, sampling methods and variational methods. Discretization is conceptually simple: we discretize every variable and then use standard discrete inference. Unfortunately, discretization requires a fine resolution for an adequate representation even of simple distributions and the situation degrades exponentially with the number of dimensions, making the approach intractable for large clique trees.

Sampling is a general method that can handle non-standard distributions such as CD CPDs; it has a low space complexity, and is guaranteed to converge as the number of samples $N$ goes to infinity. There are two main classes of sampling algorithms: those based on likelihood weighting (LW) and those based on MCMC. The advantages of LW are its generality and simplicity. However, it suffers from a few problems. First, the convergence rate is slow (on the order of $1/\sqrt{N}$). In contrast, our algorithm converges much faster in cases where the integration dimension is low. Both LW and our algorithm have problems when dealing with unlikely evidence, but the problems are much worse in LW. Continuous evidence is very problematic in LW, due to the exponential decay of the Gaussian distribution. This type of evidence has no impact on the accuracy of our algorithm. While there is some impact in the case of unlikely discrete evidence, it is much less significant than in LW; as shown in Section 6, our algorithm achieves substantially higher accuracies than LW in the same amount of running time.

On top of the slow convergence of $1/\sqrt{N}$ of sampling methods, MCMC methods converge very slowly when the mixing rate of the Markov chain is slow, which depends in unpredictable ways on the network parameters. In addition, MCMC may run into problems in arbitrary complex CD CPDs. To correctly sample a value for some variable, one has to combine all the CPDs in which it is involved into a sampling distribution — if the CPDs are complex, this task is not trivial. It would have been very interesting to compare the results of our algorithm to BUGS, but problems in BUGS prevented us from doing so. However, even our partial results imply that, at least in the case of the *Emission* network, BUGS requires many samples to achieve convergence, while our algorithm produced instantaneous answers. We do not know whether the wrong convergence is a simple implementation problem, or whether it results from a more fundamental difficulty.

Our algorithm appears closer in spirit to the variational approximation approach proposed in [11, 16]. However, the variational approach is limited to binary softmax distributions, while our algorithm is completely general. More importantly, our algorithm is guaranteed to give the correct answer (up to numerical integration errors), while the variational approach has no guarantees. It seems that variational approximation would suffer from the same problems of unlikely evidence as our approach. It is interesting to explore whether the variational approximation is sensitive to issues such as the slope of the sigmoids as is our algorithm. (I.e., is it more difficult to find a good setting for the variational parameters when the sigmoids are sharp and as a result the quality of the approximation degrades.)

The most significant limitation of our algorithm, shared with the variational approach, results from its relationship to Lauritzen's algorithm. As shown in [9], even simple CLG networks can lead to clique trees that are intractably large. An important open problem is to devise approximation schemes that are suitable for hybrid networks where Lauritzen's algorithm cannot be applied.

**Acknowledgments.** We thank Ofer Levi-Tsabari for useful discussions regarding numerical integration methods. This research was supported by ONR Young Investigator (PECASE) under grant number N00014-99-1-0464, and by ONR under the MURI program "Decision Making under Uncertainty", grant number N00014-00-1-0637. Eran Segal was also supported by a Stanford Graduate Fellowship.

## A  Parameters for Extended Crop Network

The *Policy* variable takes the values *Liberal* and *Conservative*. The *Rain* variable takes the values *Drought*, *Average*, and *Floods*. The *Profit* variable takes the values *Loss*, *Even*, and *Profit*. The *Subsidize* and *Buy* variables are both binary.

| Node | Distribution | | |
|---|---|---|---|
| Policy | | | (0.5, 0.5) |
| Rain | | | (0.35, 0.6, 0.05) |
| Subsidize | Drought | Liberal | (0.4, 0.6) |
| | Drought | Conservative | (0.3, 0.7) |
| | Average | Liberal | (0.95, 0.05) |
| | Average | Conservative | (0.95, 0.05) |
| | Floods | Liberal | (0.5, 0.5) |
| | Floods | Conservative | (0.2, 0.8) |
| Crop | Drought | | $\mathcal{N}(3, 0.5)$ |
| | Average | | $\mathcal{N}(5, 1)$ |
| | Floods | | $\mathcal{N}(2, 0.25)$ |
| Price | Yes | | $\mathcal{N}(9-C, 1)$ |
| | No | | $\mathcal{N}(12-C, 1)$ |
| Buy | | | b=-1, w=7 |
| Profit | Sub=Yes | Buy=Yes | $f_l$=exp(13-2P-PC) |
| | | | $f_p$=exp(3P+PC-23) |
| | Sub=Yes | Buy=No | $f_l$=exp(13-2P) |
| | | | $f_p$=exp(3P-23) |
| | Sub=No | Buy=Yes | $f_l$=exp(13-PC) |
| | | | $f_p$=exp(PC-23) |
| | Sub=No | Buy=No | $f_l$=exp(13) |
| | | | $f_p$=exp(-23) |

Where $P(Profit=Loss) = \frac{f_l}{f_l+1+f_p}$, $P(Profit=Even) = \frac{1}{f_l+1+f_p}$, and $P(Profit=Profit) = \frac{f_p}{f_l+1+f_p}$.

## References

[1] Y. Bar-Shalom and T. E. Fortmann. *Tracking and Data Association*. Academic Press, 1988.

[2] P. J. Davis and P. Rabinovitz. *Methods of Numerical Integration*. Academic Press, 1984.

[3] W.R. Gilks, A. Thomas, and D.J. Spiegelhalter. A language and program for complex Bayesian modelling. *The Statistician*, 43:169–78, 1994.

[4] D. Koller, U. Lerner, and D. Angelov. A general algorithm for approximate inference and its application to hybrid Bayes nets. In *Proc. UAI*, pages 324–333, 1999.

[5] A. Kozlov and D. Koller. Nonuniform dynamic discretization in hybrid networks. In *Proc. UAI*, pages 314–325, 1997.

[6] S.L. Lauritzen. Propagation of probabilities, means, and variances in mixed graphical association models. *JASA*, 87(420):1089–1108, 1992.

[7] S. Lauritzen and F. Jensen. Stable local compuation with conditional Gaussian distributions. Technical Report R-99-2014, Dept. Math. Sciences, Aalborg Univ., 1999.

[8] S.L. Lauritzen and D.J. Spiegelhalter. Local computations with probabilities on graphical structures and their application to expert systems. *J. Roy. Stat. Soc.*, B 50, 1988.

[9] U. Lerner and R. Parr. Inference in hybrid networks: Theoretical limits and practical algorithms. In *Proc. UAI*, 2001.

[10] U. Lerner, R. Parr, D. Koller, and G. Biswas. Bayesian fault detection and diagnosis in dynamic systems. In *Proc. AAAI*, pages 531–537, 2000.

[11] K. Murphy. A variational approximation for Bayesian networks with discrete and continuous latent variables. In *Proc. UAI*, pages 467–475, 1999.

[12] R Neal. Probabilistic inference using Markov Chain Monte Carlo methods. Technical Report CRG-TR-93-1, University of Toronto, 1993.

[13] V. Pavlovic, J. Rehg, T.-J. Cham, and K. Murphy. A dynamic Bayesian network approach to figure tracking using learned dynamical models. In *Proc. ICCV*, 1999.

[14] W. Press, S. Teukolsky, W. Vetterling, and B. Flannery. *Numerical Recipes in C*. Cambridge University Press, 1988.

[15] R. Shacter and M. Peot. Simulation approaches to general probabilistic inference on belief networks. In *Proc. UAI*, pages 221–230, 1990.

[16] W. Wiegerinck. Variational approximations between mean field theory and the junction tree algorithm. In *Proc. UAI*, pages 626–636, 2000.